\documentclass[10pt,twocolumn,letterpaper]{article}

\usepackage{iccv}
\usepackage{times}
\usepackage{epsfig}
\usepackage{graphicx}
\usepackage{amsmath}
\usepackage{amssymb}
\usepackage{color}
\usepackage{tabularx}
\usepackage{pifont}

\def\iccvPaperID{40} 

\ificcvfinal\fi

\iccvfinalcopy 

\begin{document}

\title{Using Image Priors to Improve Scene Understanding}

\author{Brigit Schroeder\\
University of California, Santa Cruz\\
Intel AI Lab\\
{\tt\small brschroe@ucsc.edu}
\and 
Hanlin Tang\\
Intel AI Lab\\
{\tt\small hanlin.tang@intel.com}
\and
Alexandre Alahi\\
Visual Intelligence for Transportation Lab, EPFL\\
{\tt\small alexandre.alahi@epfl.ch}
}

\maketitle

\begin{abstract}
Semantic segmentation algorithms that can robustly segment objects across multiple camera viewpoints are crucial for assuring navigation and safety in emerging applications such as autonomous driving. Existing algorithms treat each image in isolation, but autonomous vehicles often revisit the same locations or maintain information from the immediate past. We propose a simple yet effective method for leveraging these image priors to improve semantic segmentation of images from sequential driving datasets. We examine several methods to fuse these temporal scene priors, and introduce a prior fusion network that is able to learn how to transfer this information. The prior fusion model improves the accuracy over the non-prior baseline from 69.1\% to 73.3\% for dynamic classes, and from 88.2\% to 89.1\% for static classes. Compared to models such as FCN-8, our prior method achieves the same accuracy with $5\times$ fewer parameters. We used a simple encoder decoder backbone, but this general prior fusion method could be applied to more complex semantic segmentation backbones. We also discuss how structured representations of scenes in the form a scene graph could be leveraged as priors to further improve scene understanding.

\end{abstract}

\section{Introduction}

\begin{figure}[h]
\begin{center}
\includegraphics[width=.4\textwidth]{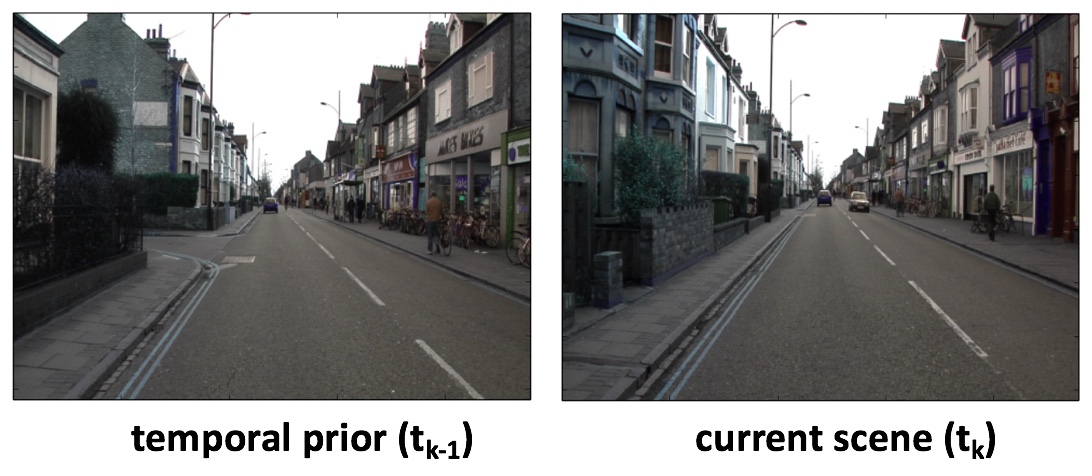}
\caption{\textbf{Scene Prior.} The image on the left is a temporal prior (one second previous) to the image on the right (representing the current timestep and scene). } 
\end{center}\label{fig:scene-prior}
\end{figure}

An autonomous vehicle is typically outfitted with several sensor modalities which can be used for mapping the environment~\cite{WanYCLZWS18} through which it drives (e.g Google self-driving cars continuously map the campus and streets of Mountain View, CA)~\cite{boudette_2017}. Image data is often collected and stored as visual maps during these traversals, including often-revisited areas such as an intersection. These scene \textit{priors}, in the form of temporal video frames, can be incorporated into scene understanding algorithms to improve the semantic segmentation of the scene. Earlier frames captured from a moving vehicle, within a time window of the current scene, often share a high degree of visual coherence (especially for objects in the distance) which can be leveraged in scene understanding algorithms. As seen in Figure \ref{fig:scene-prior}, the image on the left provides a strong prior spatially: the scene need not have the exact appearance to be useful as the fundamental layout of the road, sidewalk and buildings represent a strong structural prior.  Both recorded and live video (e.g. data previous to the current scene) provide a rich temporal prior and are a source of often unleveraged data that can enhance scene understanding.

Modeling a prior is a challenging task. Some objects, such as cars and pedestrians, are mobile and are not in the same location between frames (or time steps). The appearance of the scene can shift slightly, depending on the speed of the objects. Therefore, it can be difficult to discern which semantic labels to propagate from the prior to accurately inform the current scene. Naive approaches for selection, such as estimating the motion shift between frames, are more useful for static scenes.  Here we use a learned module to determine from raw driving data how to propagate information from the prior. 

\vspace{-3mm}

There are existing video-based approaches for semantic segmentation. Some are limited to only segmenting a limited number of objects per scene \cite{ValipourSJR17}. In \cite{ClockNets}, clockwork convnets are used for video segmentation, however segmentation is done and evaluated on limited regions of a scene (e.g. single car). A fully convolutional network (FCN-8 \cite{Shelhamer2017}) is used, which has a significantly larger number of parameters than our proposed approach (see Table \ref{table:results}). There is existing work which relies upon optical flow at inference time to warp semantic segmentation labels from the previous frame to the current, thereby  increasing the network complexity and computational cost. For example, in \cite{NilssonS18}, FlowNet\cite{FlowNet} is incorporated as an input branch of the model, significantly increasing then number of parameters in the model. NetWarp \cite{Gadde2017SemanticVC} also follows a similar approach and fuses higher-level features from a previous frame that have been warped using optical flow. We show in our preliminary approach that we are able to do full scene segmentation with multiple frames using a low complexity model. Our method is an effective but simple learned feature fusion scheme that does not require optical flow an inference. We apply this method to a basic encoder-decoder framework, but it could potentially be adapted to other more complex semantic segmentation networks. We also discuss proposed future work in which structured data in the form of a scene graph can be derived from a prior and incorporated as an additional source of a prior knowledge.

\section{Methodology}

\subsection{Prior Fusion Network Architecture}
We use a fully-convolutional encoder-decoder architecture for semantic labeling, motivated by SegNet ~\cite{segnet2015}. These models feature a bottleneck stage, where the input image in projected to a lower dimensional representation. We hypothesize that this bottleneck representation could serve as the location for incorporating prior knowledge before the decoder network expands into a semantic representation.  We define a scene prior as an image of a given location which has been captured at an earlier time step (such as frames preceding the current frame). For our experiments, we use a prior that was captured one second earlier. Early experiments showed that using a frame too far in the past to be more detrimental than helpful as the differences in the scenes (both structurally and visually) were too high. 

\begin{figure}[h]
\begin{center}
\includegraphics[width=.5\textwidth]{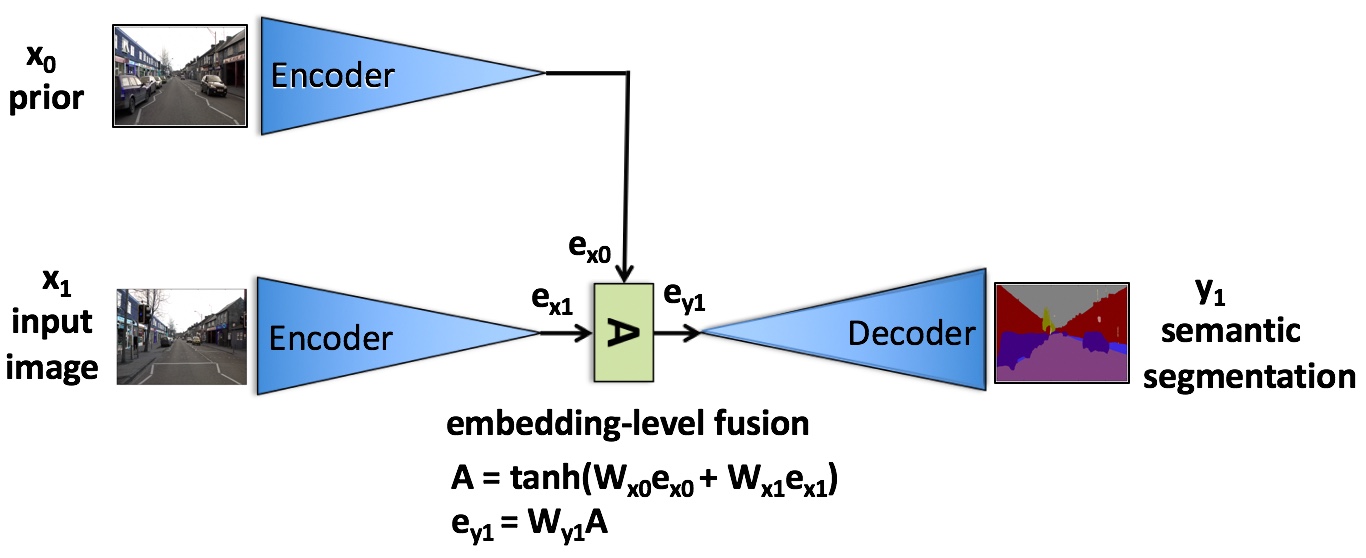}
\includegraphics[width=.5\textwidth]{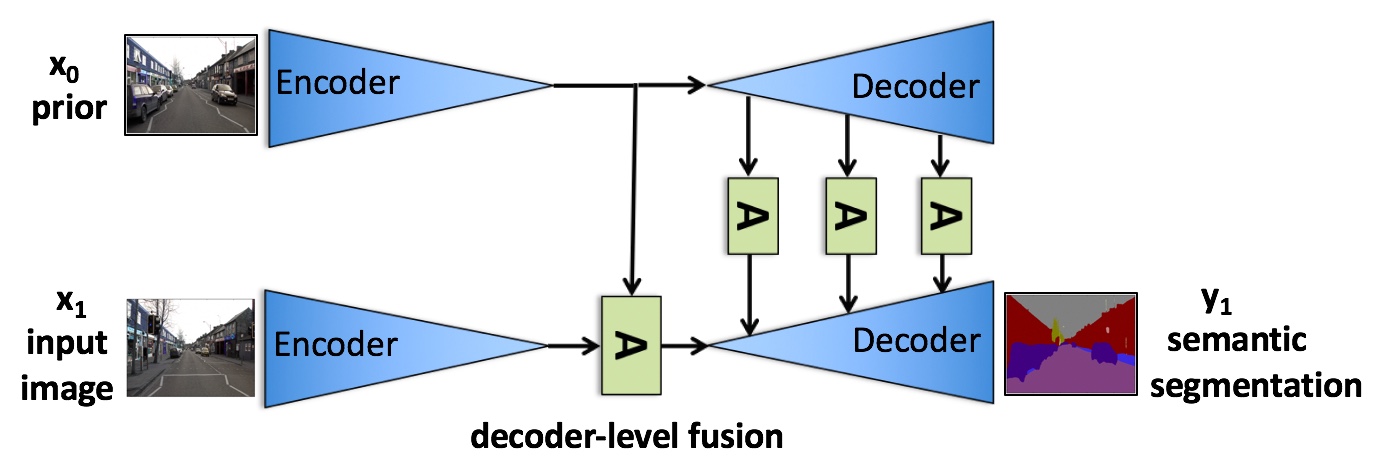}
\caption{\textbf{Multilevel Fusion Network Architectures.} A compact encoder-decoder network architecture is used with the addition embedding-level (top network architecture) and decoder-level (bottom network architecture) prior fusion. Prior features are combined using \emph{tanh} activation functions, similar to those in recurrent networks.}  \label{fig:models}
\end{center}
\end{figure}

We tested three architectures:
\begin{itemize}
\item \textbf{Baseline.} Our baseline is a fully convolutional network with eight layers. The encoder has 64-128-256-512 features, and the decoder has 512-256-128-64 features. This baseline had no access to the temporal prior.
 
 \item \textbf{Stacked Prior.} In this naive model, the prior $x_0$ and the image $x_1$ are stacked together, forming a multi-channel input, and passed into a network that is identical to the baseline.
 
 \item \textbf{Embedding Prior.} In this approach, both the prior $x_0$ and the image $x_1$ are passed through an encoder with identical structure to the baseline encoder (Figure \ref{fig:models}, top). The structure of the decoder is also identical to the baseline model and the entire model is fine-tuned from the baseline. To fuse the representations, we use a learned weighted sum with a $tanh$ activation function (module $A$ in Figure \ref{fig:models}), an idea borrowed from recurrent neural networks \cite{DonahueHRVGSD17}

\begin{equation} \label{A_eqn}
\begin{split}
& A = tanh(W_{x0}e_{x0} + W_{x1}e_{x1})\\
& e_{y1} = W_{y1} A
\end{split}
\end{equation}

$e_{x0}$ and $e_{x1}$  are the incoming features from the prior and input images to module $A$, and $e_{y0}$ are the outgoing fused features. Using this formulation, the network is able to learn how to transfer (via $W_{x0}$, $W_{x1}$ and $W_{y1}$) the prior features in the most effective way. 

 \item \textbf{Decoder Prior.} This model is similar to the embedding prior (Figure \ref{fig:models}, bottom), except the prior is applied to the decoder, and the features fused at each level of the decoder. 
\end{itemize}

Importantly, the weights for the encoder-decoder are shared between the prior network and the image network, so the only difference in parameter count between the three models above are small contributions from the $A$ modules. In early experiments, we also tested a naive approach of concatenating the features at various levels of the encoder and decoder (including bottleneck layer). While concatenation at the decoder level was more effective, the models performed poorly, so we exclude these models.

\section{Experimental Results}

\subsection{Dataset}
Each model is trained using the CamVid road scene dataset~\cite{CamVid2008} which contains several driving sequences with object class semantic labels, collected at various times of the day. There are 367 train images and 233 test images. Due to the small size of the dataset, models were initially trained with 227 x 227 random image crops from the full 360 x 480 image as in \cite{DBLP:conf/cvpr/JegouDVRB17}, and then final models were fine-tuned from these models using the full-sized images.


\begin{table}
\begin{center}
\begin{tabular}{ |l | c | c | c | c |l|} 
\hline
 & \textbf{\# Param (M)} & \textbf{Class} & \textbf{Global} & \textbf{IoU} \\
\hline
{\textbf{SegNet} \cite{segnet2015} } & 29.5 & - & 62.5& 46.4 \\
\hline
{\textbf{FCN-8}} \cite{Shelhamer2017} & 134.5 & - & 88.1 & 57.0 \\
\hline
\hline
{\textbf{BL Enc-Dec}} & 17.1 & 56.2 & 85.3 & 48.3 \\
\hline
{\textbf{Stacked Prior}} & 17.1 & 56.3 & 86.0 & 48.7 \\
\hline
{\textbf{Embed Prior}} & 24.2 & 60.5 & 88.0 & 53.7 \\ 
\hline
{\textbf{Decoder Prior}} & 26.5 &  \textbf{64.7} &  \textbf{88.9}& \textbf{57.0}\\
\hline
\end{tabular}\\
\end{center}
\vspace{1mm}
\caption{\textbf{Semantic Segmentation Evaluation.} Performance of four semantic segmentation network variants on the CamVid test set (with no class balancing), evaluated using global pixel accuracy, class accuracy and intersection over union. For model descriptions, see Section 2.}\label{table:results}
\end{table}

\begin{figure*}[h]
\begin{center}
\includegraphics[width=\textwidth]{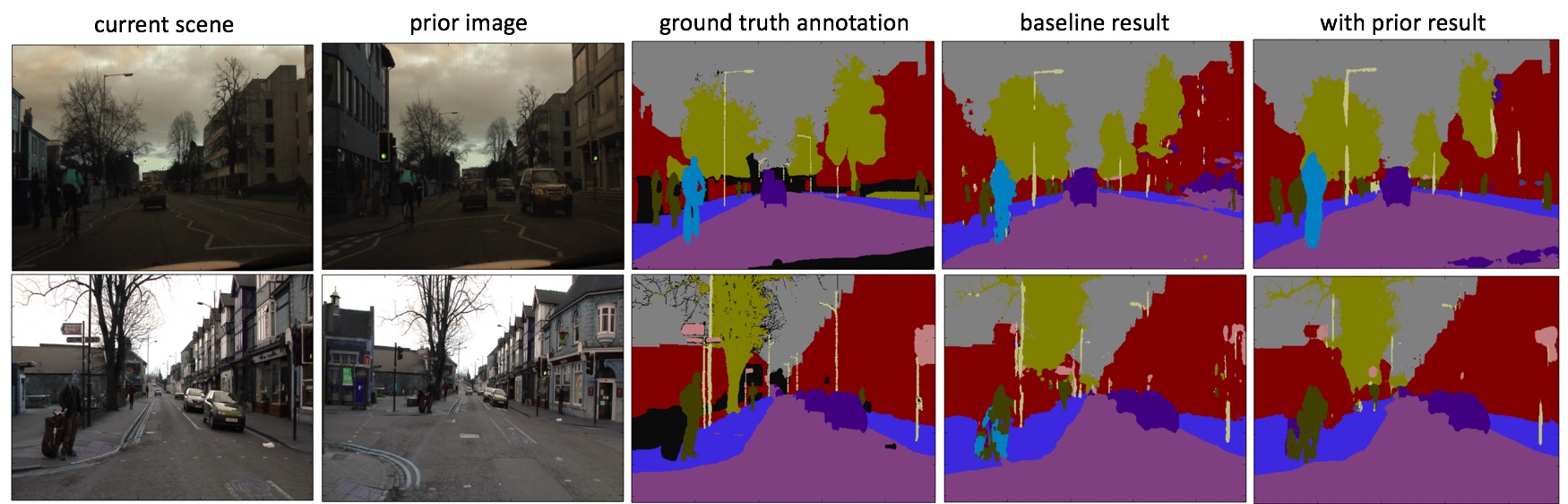}

\caption{\textbf{Semantic Segmentation with Priors.} Qualitative comparison of semantically labeled images for networks which use and do not use priors. Results from the baseline model are shown in the fourth column and Decoder Prior (our best-performing model) in the last column. } \label{table:results_pic}
\end{center}
\end{figure*}

\subsection{Prior Fusion Evaluation}
 We measured performance of scene segmentation using three standard metrics \cite{Garcia-GarciaOO17}: global accuracy, class accuracy and intersection-over-union (IoU). Global accuracy is the overall mean per-pixel labeling accuracy and class accuracy is the mean class-wise accuracy. Intersection-over-union is the average of the intersection of the prediction and ground truth regions over the union of them. As shown in Table \ref{table:results}, models that incorporate priors (Decoder Prior and Embedding Prior) significantly outperform the baseline across all three metrics in all cases. Even in the most basic prior model using stacked inputs, we see a nominal improvement over the baseline model. 

 Naively using a prior at the network input level is the least effective approach, yielding less than a  +1.0\% increase in all metrics. However, prior fusion improves most upon all metrics when done at the decoder level over the baseline. The global accuracy of per-pixel labeling increased both for fusion models, by more than +3.6\% in the best performing model. Embedding prior fusion contributes to an overall increase in global accuracy, class accuracy and IoU (in the best case, an increase of +5.4\% for IoU), indicating that fusing at the bottleneck layer alone, which has rich semantic (yet coarse) features, has significant impact on both fine-grained and coarser feature classes.  Importantly, when priors are fused at different feature resolutions throughout the decoder, all metrics further increase above the embedding prior, especially for class accuracy and IoU (+4.2\% and +3.3\%, respectively). Since higher frequency classes such as sky, building, road, etc. tend to dominate class accuracy \cite{segnet2015}\cite{Eigen2015}, this result suggests that the fusion models may increase the accuracy of object classes with low frequency, such as pedestrians, bicyclists, street signs, etc. Our models have not been pretrained on large datasets such as ImageNet \cite{imagenet}  and could most likely benefit in performance from such pretraining \cite{DBLP:conf/cvpr/JegouDVRB17}.

\begin{table}
\begin{center}

\begin{tabular}{ |l | c | c |}
\hline
 & \textbf{Static Objs} & \textbf{Dynamic Objs}\\
\hline
{\textbf{Baseline (no prior)}} & 88.2 & 69.1 \\
\hline
{\textbf{Embedding Prior}} & +0.7\% & \textbf{+4.2\%}\\ 
\hline
{\textbf{Decoder Prior}} & \textbf{+0.9\%} & +3.8\%\\
\hline
\end{tabular}\\
\vspace{2.0mm}
\caption{\textbf{Global Accuracy of Static and Dynamic Classes.} Comparison of classes which are divided into static objects (e.g. buildings, roads, trees, etc.) and dynamic objects (e.g. pedestrians, car, bicycles, etc.). The addition of a prior increases the accuracy of both types of objects.}\label{table:dynamic}
\end{center} 
\end{table}

As seen in the top half of Table \ref{table:results}, we outperform the original SegNet \cite{segnet2015}, which uses a larger number of parameters, in both mean IoU and global accuracy significantly. We achieve this by modifying the original SegNet architecture to reduce parameter count and incorporate priors,  We also match the widely-used FCN-8 model \cite{Shelhamer2017} in performance, while using $5\times$ fewer parameters. While our model is not state-of-the-art, the significant performance increases we see hold promise that our method could be applied to other types of architectures.

The performance improvement from incorporating priors is significantly enhanced when we examine dynamic versus static objects. We divided the CamVid object classes into static objects (e.g. buildings, roads, light posts, signs, trees, etc.) and dynamic objects (e.g. pedestrians, cyclists, cars, etc.). The global accuracy for both for is reported in Table \ref{table:dynamic}. Importantly, we observed that the priors had a significant impact on the semantic segmentation of dynamic objects (73.3\% versus 69.1\%), which is consistent with our expectation that objects which tend to be of smaller size, lower frequency, and significant pixel translations, would have improved accuracy with our temporal prior.

Overall, priors decrease the spurious semantic labeling of pixels, which can be seen in Figure \ref{table:results_pic}. Note that the prior model (fifth column) reduces a lot of noise in the pixel labeling and improves the labelling for fine-grained feature classes such as pedestrian and street sign, when compared to the baseline model.

\begin{figure}[h]
\begin{center}
\includegraphics[width=.50\textwidth]{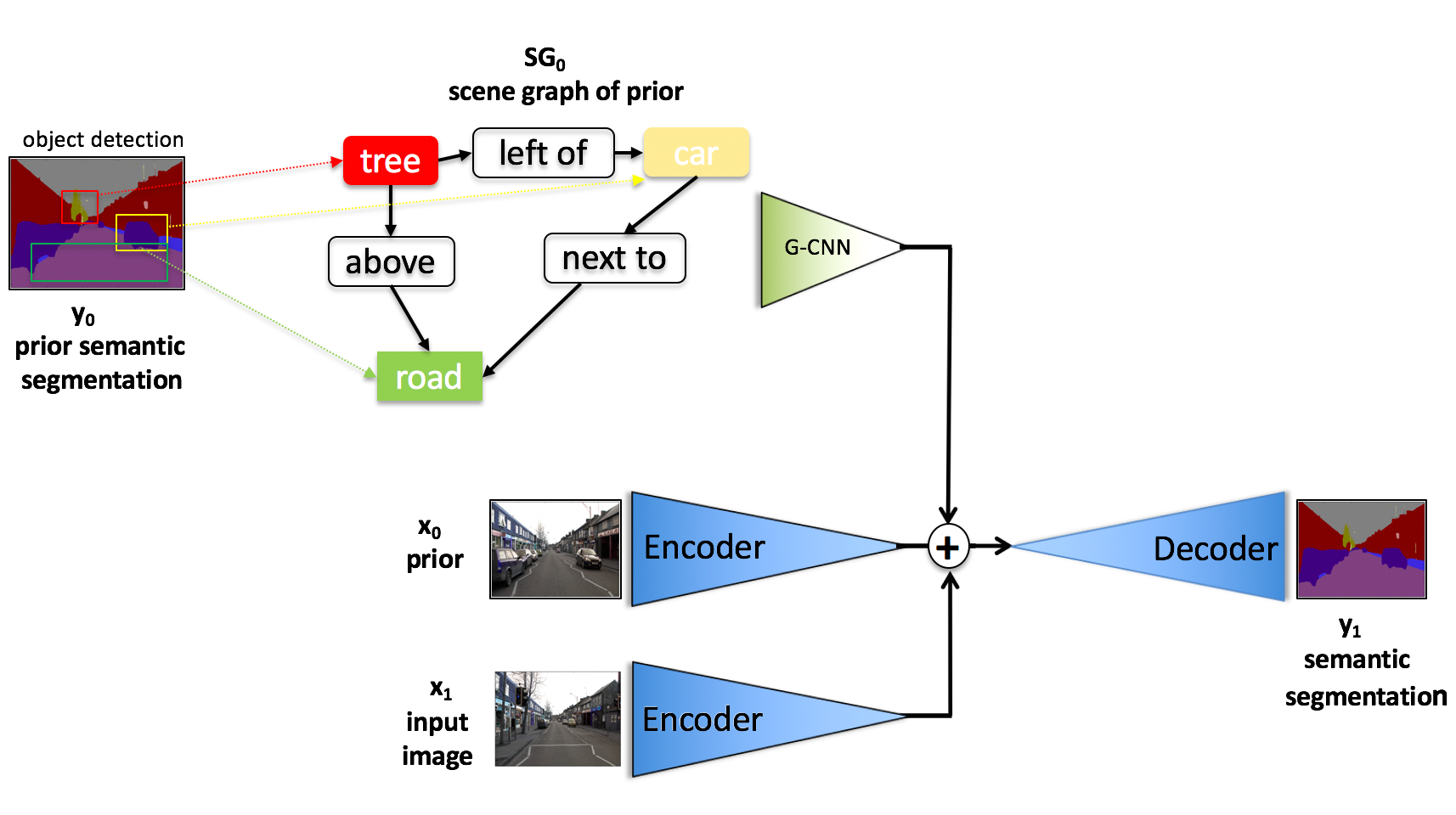}
\caption{\textbf{Prior Fusion with Scene Graphs.} A generic fusion scheme for incorporating scene graphs, a semantically-rich structured representation of scene, into a prior fusion network. A graph CNN (G-CNN) can be used to convert the scene graph into a learned feature representation.}  \label{fig:sg}
\end{center}
\end{figure}

\section{Future Work}
In future work, we plan to examine how other representations of the scene, such as scene graphs, could be provided as $structured$ prior knowledge for the network to exploit. Scene graphs encode semantic relationships between objects in an image, such as "car, next to, road" and , as illustrated in Figure \ref{fig:sg}. There we show a generic scheme for incorporating scene graphs from the prior, where objects bounding boxes from the semantic segmentation prior are structured as a scene graph (similar to how synthetic scene graphs are generated with the COCO dataset for image synthesis \cite{johnson2018image}). This structured data could then be passed through a graph convolutional neural network (G-CNN), which passes information along graph edges \cite{johnson2018image}), to produce an embedding representation of the scene. As part of future research, we seek to determine the best fusion scheme for these features, beyond incorporating them at the bottleneck layer. We also plan to examine how the output of the G-CNN network can be leveraged for relocalization purposes during navigation tasks. The scene graph embedding is a low-dimensional representation of a scene which can be searched and indexed efficiently in a stored map with key frames, such as those used in SLAM. We are motivated by \cite{Yang2018VisualSN}, which incorporates semantic knowledge, derived from knowledge graphs, as a prior in a reinforcement learning human navigation task. This leads us to believe that scene graphs would be effective form of prior knowledge to improve as semantic segmentation and potentially other scene understanding algorithms.

Unlike previous work \cite{Gadde2017SemanticVC} \cite{NilssonS18}, we propose only using optical flow at training time. We can use the optical flow to "flow" (project) the per-pixel semantic segmentation prediction from the prior to the current input image; this gives us an approximation of ground truth for the input image. We can incorporate this data to train our model in an unsupervised way, where flowed ground truth can be used in an auxiliary cross-entropy loss for predicting the class labels in the input image. Time step processing for prior and input images can also be bilateral, where prior becomes input and input becomes prior.

We have demonstrated our method on a generic encoder-decoder backbone model, however we believe this could be extended to other backbone modes such as an FCN-DenseNet. Perceptual loss \cite{JohnsonAF16}, which uses the $L_2$ differences between high-level image feature representations (rather than regular pixel values), could be incorporated into the model using the prior and input image as inputs.

\section{Conclusions}
We have demonstrated that the addition of prior knowledge to a deep convolutional network can increase the performance of semantic segmentation, particularly for dynamic objects. We introduce a method using a learned fusion module to incorporate prior information, and demonstrate that fusing at multiple feature resolutions improves performance. Our model outperforms several models of comparable size and structure and is effective in increasing the accuracy of low frequency classes. This general fusion technique could also be applied to other models beyond the encoder-decoder model architecture.

{\small
\bibliographystyle{ieee}
\bibliography{egpaper_priors}

\begin{thebibliography}{10}

\bibitem{WanYCLZWS18}
Guowei Wan, Xiaolong Yang, Renlan Cai, Hao Li, Yao Zhou, Hao Wang, and Shiyu
  Song.
\newblock Robust and precise vehicle localization based on multi-sensor fusion
  in diverse city scenes.
\newblock In {\em 2018 {IEEE} International Conference on Robotics and
  Automation, {ICRA}}.

\bibitem{boudette_2017}
Neil~E Boudette.
\newblock Building a road map for the self-driving car.
\newblock {\em The New York Times}, Mar 2017.

\bibitem{ValipourSJR17}
Sepehr Valipour, Mennatullah Siam, Martin J{\"{a}}gersand, and Nilanjan Ray.
\newblock Recurrent fully convolutional networks for video segmentation.
\newblock In {\em 2017 {IEEE} Winter Conference on Applications of Computer
  Vision, {WACV}}, pages 29--36, 2017.

\bibitem{ClockNets}
Evan Shelhamer, Kate Rakelly, Judy Hoffman, and Trevor Darrell.
\newblock Clockwork convnets for video semantic segmentation.
\newblock In {\em ECCV Workshops}, 2016.

\bibitem{Shelhamer2017}
Evan Shelhamer, Jonathan Long, and Trevor Darrell.
\newblock Fully convolutional networks for semantic segmentation.
\newblock {\em IEEE Trans. Pattern Anal. Mach. Intell.}, 39(4):640--651, April
  2017.

\bibitem{NilssonS18}
David Nilsson and Cristian Sminchisescu.
\newblock Semantic video segmentation by gated recurrent flow propagation.
\newblock In {\em 2018 Computer Vision and Pattern Recognition}.

\bibitem{FlowNet}
Alexey Dosovitskiy, Philipp Fischery, Eddy Ilg, Philip Hausser, Caner Hazirbas,
  Vladimir Golkov, Patrick van~der Smagt, Daniel Cremers, and Thomas Brox.
\newblock Flownet: Learning optical flow with convolutional networks.
\newblock In {\em Proceedings of the 2015 IEEE International Conference on
  Computer Vision (ICCV)}, ICCV '15, pages 2758--2766, Washington, DC, USA,
  2015. IEEE Computer Society.

\bibitem{Gadde2017SemanticVC}
Raghudeep Gadde, Varun Jampani, and Peter~V. Gehler.
\newblock Semantic video cnns through representation warping.
\newblock {\em 2017 IEEE International Conference on Computer Vision (ICCV)},
  pages 4463--4472, 2017.

\bibitem{segnet2015}
Vijay Badrinarayanan, Alex Kendall, and Roberto Cipolla.
\newblock Segnet: A deep convolutional encoder-decoder architecture for image
  segmentation.
\newblock {\em IEEE Transactions on Pattern Analysis and Machine Intelligence},
  2017.

\bibitem{DonahueHRVGSD17}
Jeff~Donahue et~al.
\newblock Long-term recurrent convolutional networks for visual recognition and
  description.
\newblock {\em {IEEE} Trans. Pattern Anal. Mach. Intell.}

\bibitem{CamVid2008}
Gabriel~J. Brostow, Julien Fauqueur, and Roberto Cipolla.
\newblock Semantic object classes in video: A high-definition ground truth
  database.
\newblock {\em Pattern Recognition Letters}, 2008.

\bibitem{DBLP:conf/cvpr/JegouDVRB17}
Simon~J{\'{e}}gou et~al.
\newblock The one hundred layers tiramisu: Fully convolutional densenets for
  semantic segmentation.
\newblock In {\em 2017 {IEEE} CVPR Workshops}.

\bibitem{Garcia-GarciaOO17}
Alberto Garcia{-}Garcia, Sergio Orts{-}Escolano, Sergiu Oprea, Victor
  Villena{-}Martinez, and Jos{\'{e}}~Garc{\'{\i}}a Rodr{\'{\i}}guez.
\newblock A review on deep learning techniques applied to semantic
  segmentation.
\newblock {\em CoRR}, abs/1704.06857, 2017.

\bibitem{Eigen2015}
David Eigen and Rob Fergus.
\newblock Predicting depth, surface normals and semantic labels with a common
  multi-scale convolutional architecture.
\newblock In {\em Proceedings of the 2015 IEEE International Conference on
  Computer Vision (ICCV)}, ICCV '15, pages 2650--2658, Washington, DC, USA,
  2015. IEEE Computer Society.

\bibitem{imagenet}
J.~Deng, W.~Dong, R.~Socher, L.-J. Li, K.~Li, and L.~Fei-Fei.
\newblock {ImageNet: A Large-Scale Hierarchical Image Database}.
\newblock In {\em CVPR09}, 2009.

\bibitem{johnson2018image}
Justin Johnson, Agrim Gupta, and Li~Fei-Fei.
\newblock Image generation from scene graphs.
\newblock In {\em CVPR}, 2018.

\bibitem{Yang2018VisualSN}
Wei Yang, Xiaolong Wang, Ali Farhadi, Abhinav Gupta, and Roozbeh Mottaghi.
\newblock Visual semantic navigation using scene priors.
\newblock {\em ArXiv}, abs/1810.06543, 2018.

\bibitem{JohnsonAF16}
Justin Johnson, Alexandre Alahi, and Li~Fei{-}Fei.
\newblock Perceptual losses for real-time style transfer and super-resolution.
\newblock In {\em {ECCV} {(2)}}, volume 9906 of {\em Lecture Notes in Computer
  Science}, pages 694--711. Springer, 2016.

\end{thebibliography}
}

\end{document}